\documentclass{article}
\usepackage{ijcai21}

\usepackage{times}
\usepackage{soul}
\usepackage{url}
\usepackage[hidelinks]{hyperref}
\usepackage[utf8]{inputenc}
\usepackage[small]{caption}
\usepackage{graphicx}
\usepackage{amsmath}
\usepackage{amsthm}
\usepackage{booktabs}
\usepackage[ruled,vlined]{algorithm2e} 
\usepackage{algorithmic}
\usepackage{subcaption}
\usepackage{todonotes}
\usepackage{color}
\usepackage{multirow}

\title{Lessons from the Clustering Analysis of a Search Space: A Centroid-based Approach to Initializing NAS}

\author{
Kalifou Rene Traore$^{1,2}$
\and
Andr\'es Camero$^1$\and
Xiao Xiang Zhu$^{1,2}$
\affiliations
$^1$German Aerospace Center (DLR), Remote Sensing Technology Institute (IMF), Germany\\
$^2$Technical University of Munich, Data Science in Earth Observation, Germany\\
\emails
\{kalifou.traore, 
andres.camerounzueta, 
xiaoxiang.zhu\}@dlr.de
}

\begin{document}

\maketitle

\begin{abstract}

Lots of effort in neural architecture search (NAS) research has been dedicated to algorithmic development, aiming at designing more efficient and less costly methods.
Nonetheless, the investigation of the initialization of these techniques remain scare, and currently most NAS methodologies rely on stochastic initialization procedures, because acquiring information prior to search is costly.
However, the recent availability of NAS benchmarks have enabled low computational resources prototyping. 
In this study, we propose to accelerate a NAS algorithm using a data-driven initialization technique, leveraging the availability of NAS benchmarks.
Particularly, we proposed a two-step methodology. First, a calibrated clustering analysis of the search space is performed. Second, the centroids are extracted and used to initialize a NAS algorithm.
We tested our proposal using Aging Evolution, an evolutionary algorithm, on NAS-bench-101. The results show that, compared to a random initialization, a faster convergence and a better performance of the final solution is achieved.

\end{abstract}

\section{Introduction}




Recently, significant effort has been made to improve the reproducibility in Neural Architecture Search (NAS), and to democratize its research~\cite{elsken2019neural}.
In this context, the release of performance evaluation databases~\cite{pmlr-v97-ying19a,dong2020nasbench201} has enabled low computational cost algorithm benchmark.
However, so far, few authors have take advantage of these sources of data to improve the design of NAS algorithms.

This study sets out to answer the following question: can we improve the performance of a population based NAS algorithm by initializing its population with a data-driven approach? 
To this problem, we propose a two-step approach.
First, a tailored clustering analysis of a target search space is performed. Second, after obtaining satisfying quantitative clustering results, the centroids are extracted and used to initialize a population based NAS algorithm.

To validate our proposal we selected a state-of-the-art evolutionary algorithm, Aging Evolution~\cite{AgingEvol2019Real}, and we tested the proposed initialization on NAS-bench-101~\cite{pmlr-v97-ying19a}. 
The results show that, compared to random initialization, our approach improves the convergence and the performance of the final solution (for 36 epochs), using a compact feature representation.

The remainder of this article is as follows:
The following section introduces NAS and Aging Evolution~\cite{AgingEvol2019Real}. 
Section ~\ref{sec:methodology} describes the proposed methodology.
Section~\ref{sec:experimental-setup} introduces the experimental setup. Section ~\ref{sec:results} presents the results.
Finally, Section~\ref{sec:conclusion} outlines the conclusions.

\section{Related Work}\label{sec:related-work}

The following Section is dedicated to introducing the general topic of NAS,
as well as \emph{Aging Evolution}, one of the early evolutionary NAS algorithms reaching state-of-the-art performances on modern \emph{computer vision} (CV) benchmarks.

\subsection{Neural Architecture Search}
NAS is a subfield of Automated Machine Learning (AutoML)~\cite{hutter2019automated}. 
It aims at finding architectures providing the best performances on unseen data for a given task to learn and provided data set~\cite{elsken2019neural}.
NAS has seen successes in various CV tasks including image classification and object  detection; but also other areas of application such as Natural Language Processing (NLP). 
Early works on NAS were using Evolutionary-based approaches to optimize the model (i.e., the architecture, weights, and/or nodes)~\cite{Ojha2017}. More recently, the scope of the search algorithms used has been extended, including such as Bayesian Optimization~\cite{camero2021bayesian} and Differentiable Search Algorithms~\cite{DARTS}.

\subsection{Aging Evolution}
In recent years, an \textit{Evolutionary Algorithm} based approach to NAS achieved state-of-the-art performance on classical CV benchmarks: the \emph{Aging Evolution}\cite{AgingEvol2019Real}, outlined in Algorithm~\ref{algo:aging-evol}.

This particular baseline
\footnote{The implementation of \emph{Aging Evolution} is available on NASBench-101 repository.} 
evolves a population of constant size~$P$ and returns a highest performing solution after $C$ generations. 
The performance of each candidate solution is evaluated using the \emph{fitness} function (Train\_and\_eval). Particularly, the fitness is computed by training a solution using Stochastic Gradient Descent (SGD), and evaluating its classification accuracy (Top-1 in validation and test).
First, a population is randomly initialized and its fitness is computed.
Then, the population is evolved until the number of evaluated generations is greater than $C$, by a tournament selection, mutate operation, and aging replacement.
The \textit{tournament selection} of size $S$ consists of randomly selecting $S$ solutions (with replacement) from the current population. Then, the highest performance solution is selected (\emph{parent}).
Later, the \emph{parent} solution is mutated (\emph{child.arch}) by a two step process: A \textit{hidden state mutation}, the connections between operations in a graph-represented solution (cell) are modified, and an \textit{operation mutation}, the operation within the \textit{cell} is modified.
Then, the \emph{child.arch} is evaluated.
Finally, the \emph{oldest} solution of the population (i.e., the earliest evaluated solution in the population) is replaced by the new candidate solution (\emph{child.arch}).
The authors of Agign Evolution claim that exists a parallel between the introduced age-based removal to a \textit{regularization} of the evolution. Once the termination criteria is met, the best solution of the population is returned.

\vspace{-0.2cm}
\begin{algorithm}[h]
\footnotesize
\SetAlgoLined
$population \gets$ empty queue\\
$history \gets \emptyset$\\

\While{$\mid population\mid < P$ }{
    $model.arch \gets$ Random\_Architecture()\\
    $model.accuracy \gets$ Train\_and\_Eval($model.arch$)\\
    add (right) $model$ to $population$\\
    add $model$ to $history$\\
}
\While{$\mid history\mid < C$ }{ 
    $sample \gets \emptyset$\\
    \While{$\mid sample\mid < S$ }{
        $candidate \gets$ random\_element from $population$\\
        add $candidate$ to $sample$\\
    }
    $parent \gets$ highest\_accuracy model $\in sample$\\
    $child.arch \gets$ Mutate($parent.arch$)\\
    $child.accuracy \gets$ Train\_and\_Eval($child.arch$)\\
    add (right) $child.arch$ to $population$\\
    add $child$ to $history$\\
    remove (left) $dead$ from $population$\\
    discard $dead$\\
}
return highest-accuracy $model$ in $history$\\
\caption{Aging Evolution}
\label{algo:aging-evol}
\end{algorithm}
\vspace{-0.2cm}

A recent review~\cite{Kazimipour2014Review_EA_Init}  on initialization techniques to Evolutionary Algorithms 
describes existing methods with three characteristics: \textit{Randomness}, \textit{Compositionality} and \textit{Generality}. 
In the case of the \textit{Aging Evolution} baseline, its initialization relies on  \textit{Randomness}, i.e $P$ samples 'drawn uniformly at random'. Therefore, it is a non-composite and generic procedure. 


\section{A Data-driven approach to initializing a NAS Search Strategy} \label{sec:methodology}

This section introduces the proposed approach of Cluster Analysis for enhancing the performances of a NAS algorithm. 
First, we describe the overall pipeline of the methodology.
Second, we detail the feature engineering essential to the analysis.

\subsection{Pipeline}

This study aims at leveraging the knowledge about a Search Space to help improve 
the performances of a Search Strategy.
In particular, it sets sets out to answer the following question:
can we improve the convergence of a population-based NAS algorithm 
by initializing it with a data-driven approach?

\begin{figure}[ht]
  \includegraphics[width=0.99\linewidth]{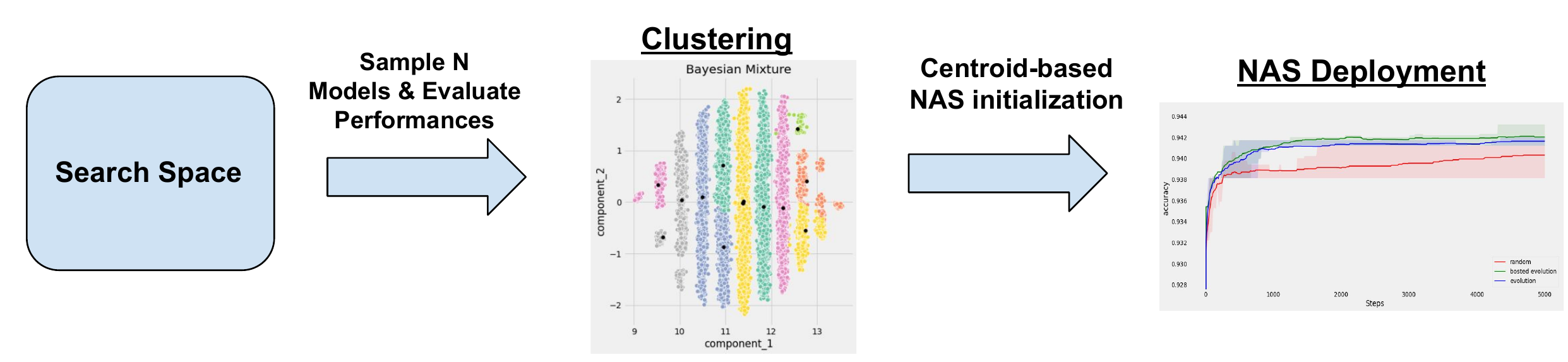}\hfill
  \caption{ The pipeline of the proposed data-driven approach to NAS initialization. }
  \label{fig:pipeline}
\end{figure}

To tackle this problem, we propose an approach consisting of two steps, depicted in Figure~\ref{fig:pipeline}.
First, we perform a performance-based clustering analysis of the Search Space.
Given the search space and a machine learning task, we sample a set of $N$ architectures. 
Each architecture is trained, evaluated, and encoded using the procedures described in Section~\ref{subsec:encoding}.
As the feature vector consists in an architecture representation and its performance in test, 
the resulting clusters should relate to specific behaviors (performances) on the learned task. 
Moreover, processing high dimensional and sparse data can sometimes be uneasy, therefore we propose to facilitate the clustering by reducing the dimension of input features.
With the reduced samples, we proceed with a clustering analysis composed itself of a sequence of sub-steps.
This sequence is as follows: 
we reduce the dimension of the samples, 
perform the clustering and assessing qualitatively and quantitatively its results.
Besides, calibrating the proper number of clusters play an important roles 
in the results retrieved. 
Thus, we seek to identify values to this hyperparameter providing satisfying results.


In second step of the proposed methodology, 
we extract the centroids obtained in the clustering, 
initialize a population-based algorithm (\textit{Aging Evolution}),
and assess it performances.   

To summarize, this approach to search initialization is: 
\begin{itemize}
    \item \textit{composite}: it is multi-step initialization procedure relying on sampling a search space,
    clustering it, and initializing an algorithm with the centroids extracted.
    \item \textit{generic}: it is not application-specific, in fact the clustering could be done on any type of search space given an encoding including a solution representation and its fitness evaluation.
    \item \textit{deterministic or stochastic}: the stochasticity of the procedure depends on the stochasticity of tool selected for clustering.
\end{itemize}



\subsection{Feature representation} \label{subsec:encoding}

To best take advantage of information about the search space when clustering, 
we first introduce a minimal feature engineering.

As we look to uncover models and structures relevant to NAS algorithms via clustering, 
we seek a feature representation encoding an architecture as well as its performances.
As in~\cite{pmlr-v97-ying19a}, we consider neural architectures identified by an elementary component repeated in blocks, a feed-forward cell.
Such cell is a DAG, with a maximum number of operations (nodes),
a maximum number of transformations (edges) and a fixed set of possible operations (e.g Max pool, convolution 3x3) labeling each node.
A cell is in practice represented a list of selected operations and an adjacency matrix of variable size.

Therefore, we construct two versions of clustering feature representation, both in the form of vectors.
The first one (\textbf{Original}, Short Encoding) consists in concatenating for each model, its adjacency matrix,
the list of operations, and the list of performances in test for all available training duration $\{t_0, t_1, t_2, t_3\}$.
Note that this is a variable length feature representation due to the nature of the adjacency matrix.

Alternatively, the second representation (\textbf{Binary}, Long Encoding) corresponds to the expanded adjacency matrix, i.e., the matrix that consider all possible operations (according to the constrains of the search space). This is a fixed length encoding. 
Moreover, for both encoding, the vector form of the adjacency matrix is obtained 
by a flattening in row-major fashion (C-style). 

\vspace{-0.2cm}
\section{Experimental Setup}\label{sec:experimental-setup}

The experiments performed aim to validate that the initialization of a population-based NAS Algorithm can benefit from models identified via Clustering Analysis of a Search Space.
In this Section, first, we introduce the problem used to validate our proposal. Second, we present the parameters used for performing the experiments on Clustering.
Third, we detail the performance metrics used to assess the quality of the clustering. 

\subsection{NASBench-101} \label{sec:nb101-dataset}

NASBench-101 is a database of neural network architectures and their performance evaluated on the data set of CIFAR-10. It contains $N=450K$ unique architectures~\cite{pmlr-v97-ying19a}.
Indeed, to tackle the given machine learning task of CIFAR-10,
all contained models use of a classical image classification structure similar to ResNets.
Indeed, the backbone of a model contains a head, a body and a tail.
Its body is made by alternating three (3) times a block with a down-sampling module.
Each block is obtained by repeating three (3) times a module called 'cell'. 
A cell is a computational unit that can be represented by a Directed Acyclic Graph (DAG).
It consist in an input node, an output node and intermediate nodes representing operations (Conv 3x3, Conv 1x1, Maxpool 3x3), and connections indicating features being transformed. Therefore, each architecture differs by its cell. 
In practice, the DAG of a model is encoded by an adjacency matrix and a list of operations labelling the associated nodes. The constrains on such DAG are the following: the can be at most $N=7$ nodes and $E=9$ edges in a cell . 

Moreover, all models were trained for 108 epochs using the same experimental setting (learning rate etc), but performance evaluations in training, validation and test were 
also provided after 4, 12 and 36 epochs.

\subsection{Hyperparameters for Clustering}

All the clustering experiments were done with a set of $N=10000$ randomly sampled models.
The considered clustering algorithms are K-means, DBSCAN, BIRCH, Spectral Clustering, and a Bayesian Gaussian Mixture model (BGM). All 
were obtained from the latest version (0.24.1) of the \textit{Scikit-learn} library ~\cite{scikit-learn}. 
Table~\ref{table:hyperparam-clustering} shows the hyperparameters selected for each, including
the maximum number of iterations (max iter), the number of samples used at initialization, 
or other more algorithm-specific ones.
Note that they are either default ($N/A$) or 
slightly modified to provide satisfying 
clustering performances. 

\begin{table}[ht]
 \centering 
 \scriptsize
\begin{tabular}{lrrr}\hline 
\textbf{Method} & \textit{max iter} & \textit{N init} &  \textit{other} \\\hline 
KMEANS & 500 & 50 & kmeans++ init \\
DBSCAN & 500 &  200 & eps=0.30 \\
BIRCH & 500 & N\/A & threshold=0.12\\
SPECTRAL & 500 & N\/A& N\/A\\
BGM  & 500 & N\/A& Dirichlet weight distribution, full co-variance\\
\hline  
\end{tabular}
\caption{Hyperparameters of the clustering algorithms.}
\label{table:hyperparam-clustering}
\end{table}

\vspace{-0.4cm}
\subsection{Clustering performance evaluation}\label{subsec:clustering-metrics}

Moreover, we use various ways of assessing the quality of the results for each step of the approach. 
Regarding step one (1), we propose to measure the Clustering performance using the following three (3) standard metrics: the Silhouette Coefficient~\cite{SilhouetteCoefficient}, the Calinski-Harabasz~\cite{CalinskyHarabaszIndex} and the  Davies-Bouldin Indexes~\cite{DaviesBouldinIndex}.
These inform on how well separated and dense are the resulting clusters.
They all apply in the context of Clustering with missing labels, which is relevant as we 
seek to investigate relevant clusters and features for NAS algorithms without prior assumptions. 
The Silhouette Coefficient is a metric comprised between -1 and +1, with higher values associated to more dense and separated clusters
The Calinski-Harabasz Index also rates a better defined clusters with higher values.
Similarly, the Davies-Bouldin Index, measures a 'similarity' between clusters, providing smaller values for better clustering.
%
%
Additionally we propose to corroborate the later with a qualitative analysis (visual assessment) for validation before for step two.

Regarding step two (2), we assess the quality of the centroid-based initialization using the performances of the algorithm (best obtained accuracy in test). More importantly, we compare its performances to those of  initializing the same algorithm with random samples, or Random Search.

\section{Results}\label{sec:results}

In this section, 
we present results on Clustering for accelerating NAS algorithms.
First, we show results on selecting the proper dimension reduction tool and hyperparameters for the Clustering.
Then, we show results on identifying the number of clusters providing satisfying Clustering performances.
We also present results on Qualitatively assessing the clusters quality for various algorithms.
Last but not least, we present results on improving NAS performances using a centroid-based initialization of a evolutionary NAS algorithm.

\subsection{Dimension Reduction}


To begin our experimental study, we seek to calibrate the dimension reduction of the input features.

\begin{figure}[ht]
  \includegraphics[width=0.99\linewidth]{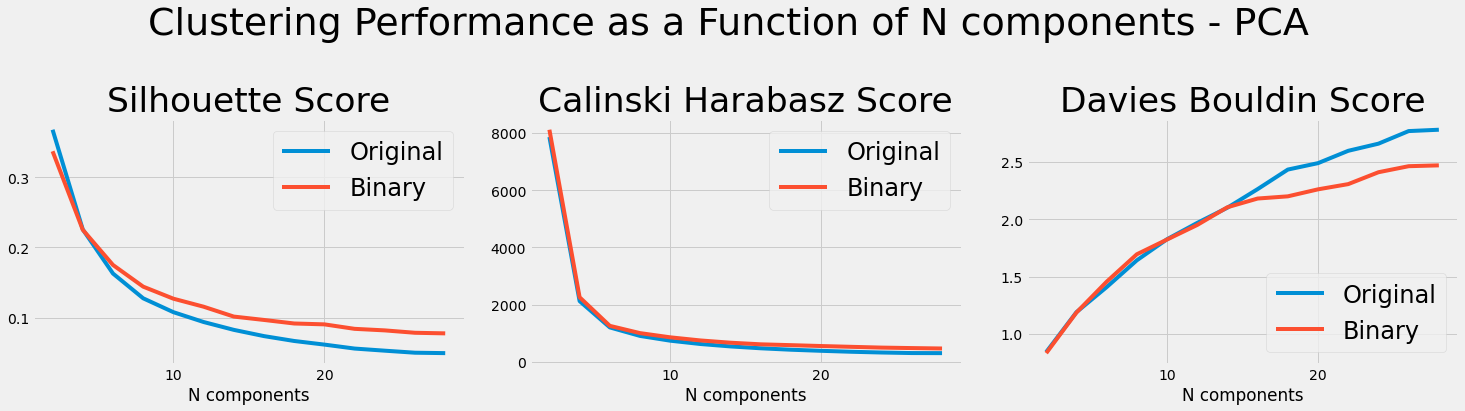}\hfill
  \caption{Input feature reduction for clustering, with an arbitrary number of clusters of $N=10$. }
  \label{fig:dim-red-pca}
\end{figure}

Figure~\ref{fig:dim-red-pca} show clustering performances as a function of the number of components of input features.
The blue and red curves displays performances using respectively the Short (Original) and the Long Encoding (Binary).
The dimension reduction is performed using PCA and clustering with K-means.

Using the Short Encoding (Original), the three metrics are in favor of using a small number 
of components for input features via PCA. Indeed the smaller the number of components the higher the Silhouette and Calinski-Harabasz scores, 
and the lower the Davies-Bouldin index, with optimal values for using two (2) components.
The same observations applies when using the Long Encoding (Binary).
    
Using the Long Encoding (Binary) yields slightly better performances then the Short Encoding (Original),
with an sensible improvement with larger number of components with PCA.

\begin{figure}[ht]
  \begin{subfigure}{\linewidth}
  \includegraphics[width=0.99\linewidth]{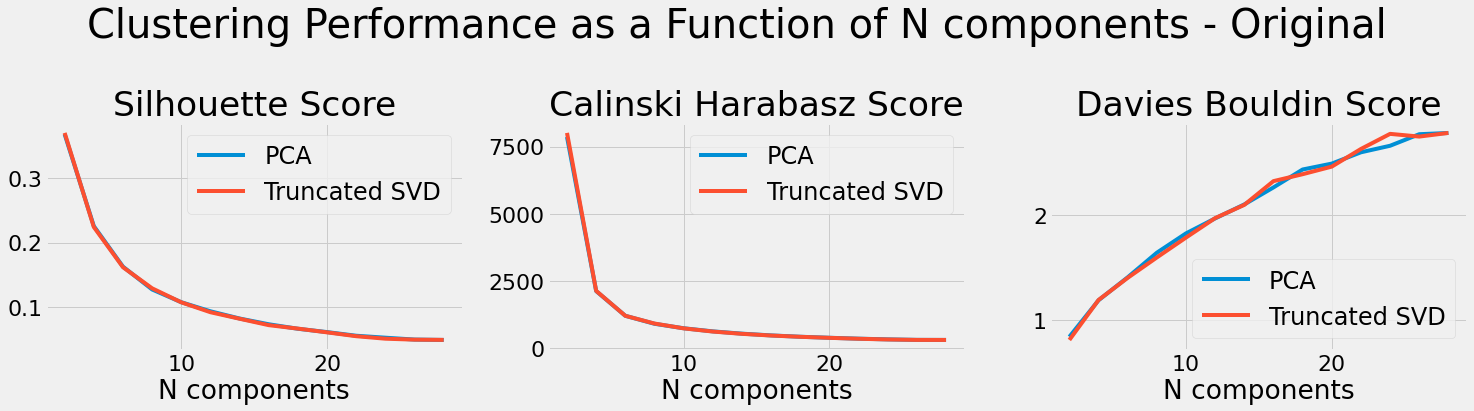}\hfill
  \caption{}
  \end{subfigure}\par\medskip
  \begin{subfigure}{\linewidth}
  \includegraphics[width=.99\linewidth]{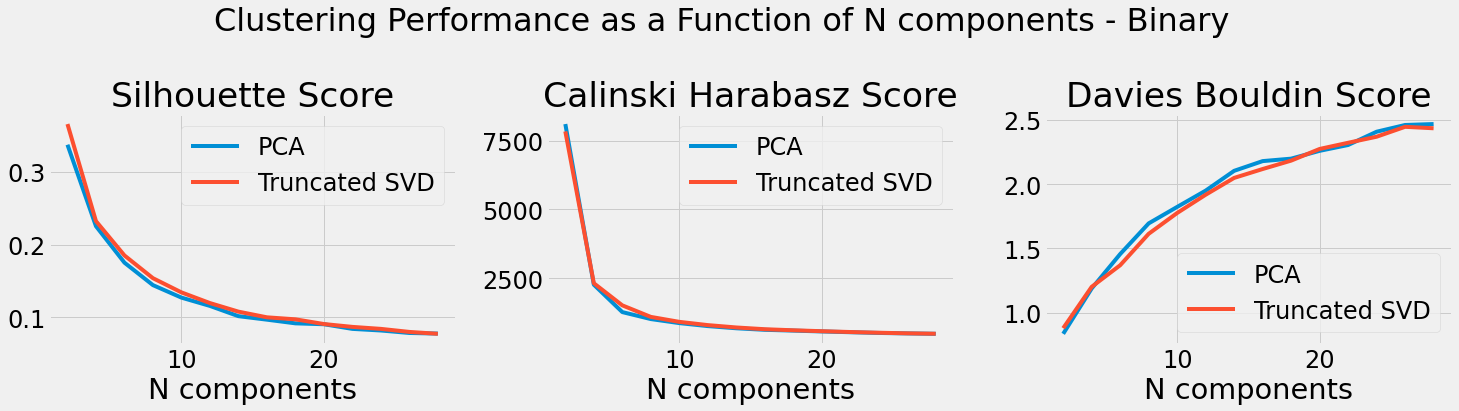}\hfill
  \caption{}
  \end{subfigure}
  \caption{Identifying the proper reduction tool for sparse data, using an arbitrary number of clusters of $N=10$.}
  \label{fig:pca-vs-truncsvd}
\end{figure}

As both encoding are rather sparse (lenght of up to 58, or up to 298), we look know if the dimension reduction tool is affected by such sparsity.
Figure~\ref{fig:pca-vs-truncsvd} show clustering performances as a function of the number of components of input features, for various reduction tool.
The blue and red curves displays performances using respectively PCA and Truncated SVD as dimension reduction tools.
Plot (a) and (b) display results using respectively the Short (Original) and the Long Encoding (Binary).
The clustering is performed with K-means.

Trying an alternative dimensional reduction tool (Truncated SVD) more suitable for higly sparse data
does not worsen results on the Short (Original) encoding (see Figure ~\ref{fig:pca-vs-truncsvd} (a)). Moreover, it allows for a slight improvement over PCA when using the Long (Binary) Encoding (see Figure ~\ref{fig:pca-vs-truncsvd} (b)).

To summarize, the findings show that reducing the dimensions of the input features to 2D provides the best performances on both encoding. Using the Long (Binary) encoding improves the results.
Also, using Truncated SVD shows slight improvements as it is more suitable for sparse data.
Given these findings, we following experiments are performed using Truncated SVD for a 2D reduction of input.

\subsection{Number of Clusters}

Next, we look to identify the number of clusters providing the most satisfying results.

\begin{figure}[ht]
  \includegraphics[width=0.99\linewidth]{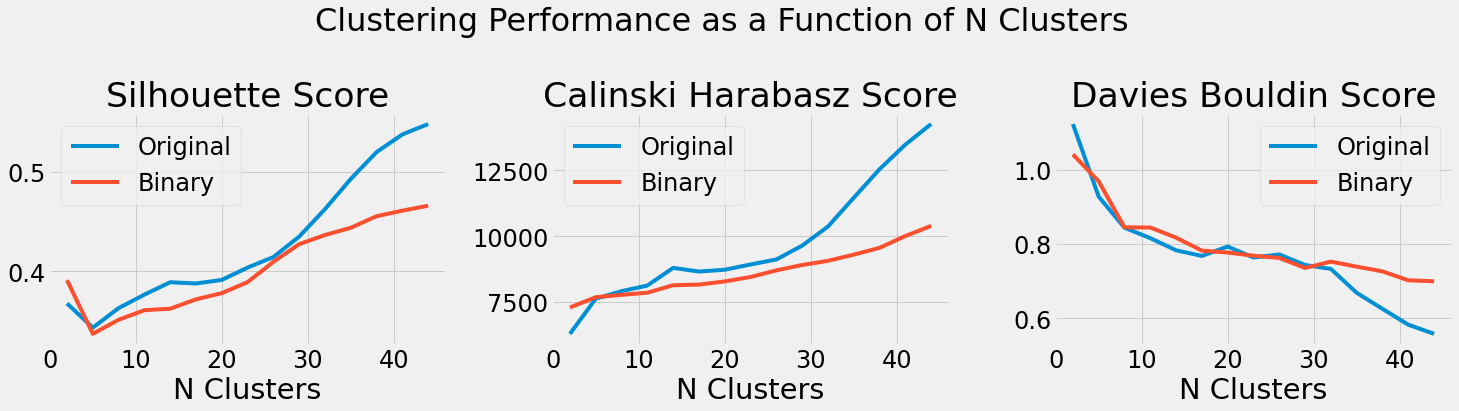}\hfill
  \caption{Identifying the proper number of clusters, using Truncated SVD for dimension reduction ($N=2$ components) and K-means.
  }
  \label{fig:n-clusters}
\end{figure}

Figure~\ref{fig:n-clusters} show clustering performances as a function of the number of clusters.
The blue and red curves displays performances using  results using respectively the Short (Original) and the Long Encoding (Binary). All input features were reduced to two (2) components using Truncated SVD, 
and clustering is performed with K-means.

Using both encoding, all performance metrics points towards the use of large number of clusters.
Indeed the higher the number of clusters, the higher the Silhouette and Calinski-Harabasz scores, and the lower the Davies-Bouldin index. 
Additionally, intermediate values around twenty (20) and twenty seven (27) clusters respectively for the Original and Binary encoding seem to reach satisfying performance already. 

Therefore, results suggest using an intermediate (20,30) to large number of clusters 
for improving the K-means clustering performances, with a preference for the Short encoding. 

\subsection{Qualitative Cluster Analysis}

As an additional way to validate the clustering results, 
we seek to visualize the obtained clusters 
and compare them to the natural layout of the reduced data.

\begin{figure}[ht]
  \begin{subfigure}{\linewidth}
  \includegraphics[width=0.99\linewidth]{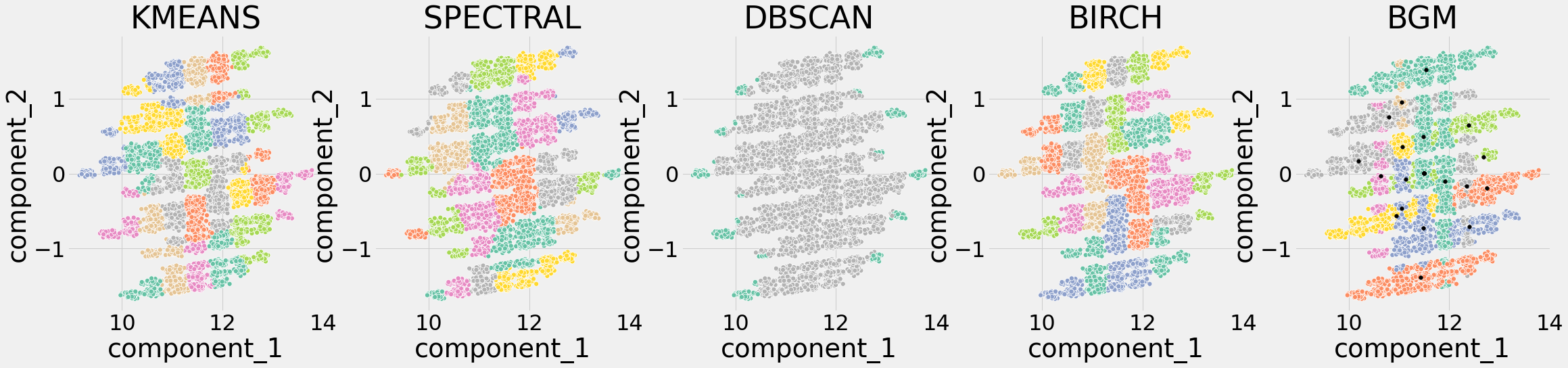}\hfill%
  \caption{}
  \end{subfigure}\par\medskip
  \begin{subfigure}{\linewidth}
  \includegraphics[width=.99\linewidth]{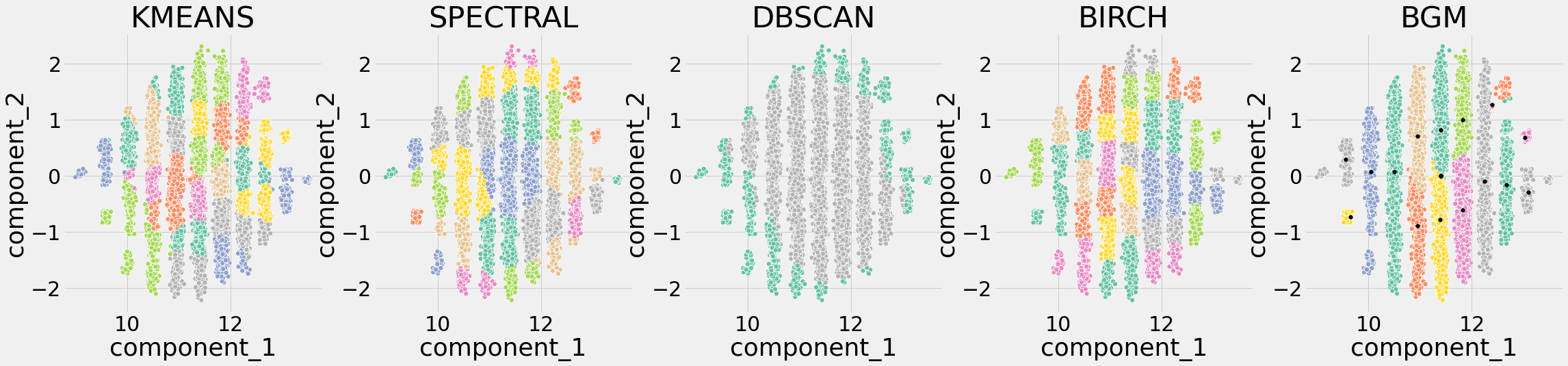}\hfill
  \caption{}
  \end{subfigure}
  \caption{Qualitative analysis of clustering for both feature representations. 
  Here we use Truncated SVD for dimension reduction, and $N=2$ components.}
  \label{fig:qualitative}
\end{figure}

Figure~\ref{fig:qualitative} displays visual clustering results for five algorithms: K-means, Spectral Clustering, DBSCAN, Birch, and a Bayesian Gaussian Mixture of models (BGM). All input features were reduced to two (2) components using Truncated SVD.
Plot (a) and (b) display results using respectively the Short (Original) and the Long Encoding (Binary).

When using the Short encoding as depicted in Figure~\ref{fig:qualitative} (a), the clusters seem to have natural horizontal to diagonal (45 degree) layout.
This layout is not well captured by the evaluated algorithms.
The Bayesian Gaussian mixture of models (BGM) seems to provide the most satisfying results, despite little calibration.

When using the Long encoding as depicted in Figure~\ref{fig:qualitative} (b),
clusters naturally layout in well separated and vertical columns.
This is also best captured by the BGM. 

Overall, results suggest using BGM for robust clustering on both feature representations.

\subsection{Towards Improving NAS Algorithms: A Centroid-based Initialization}

In order to assess the quality of the centroids extracted, 
we use them for initializing the \textit{Aging Evolution}.

\begin{figure}[]
  \begin{subfigure}{\linewidth}
  \includegraphics[width=0.99\linewidth]{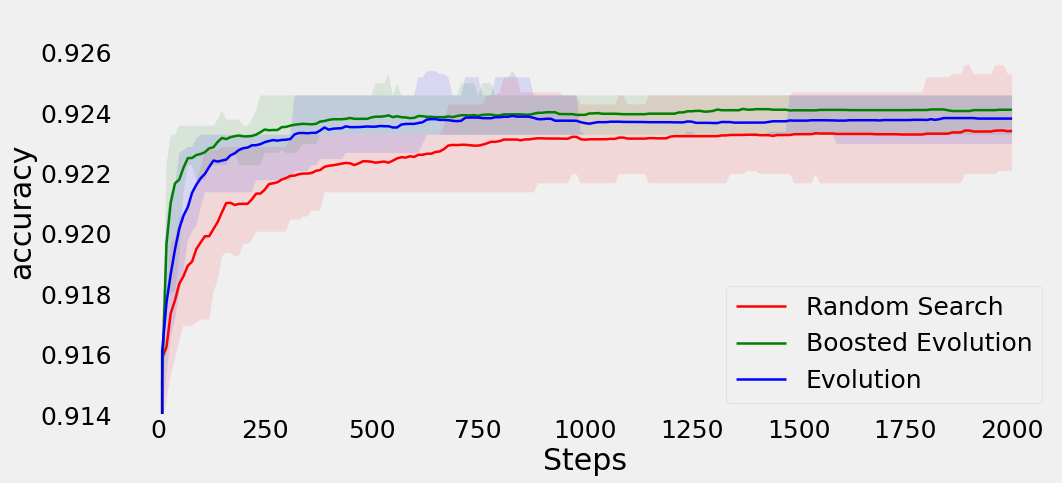}\hfill
  \caption{}
  \label{fig:boosting-ga-original-36}
  \end{subfigure}\par\medskip
  \begin{subfigure}{\linewidth}
  \includegraphics[width=.99\linewidth]{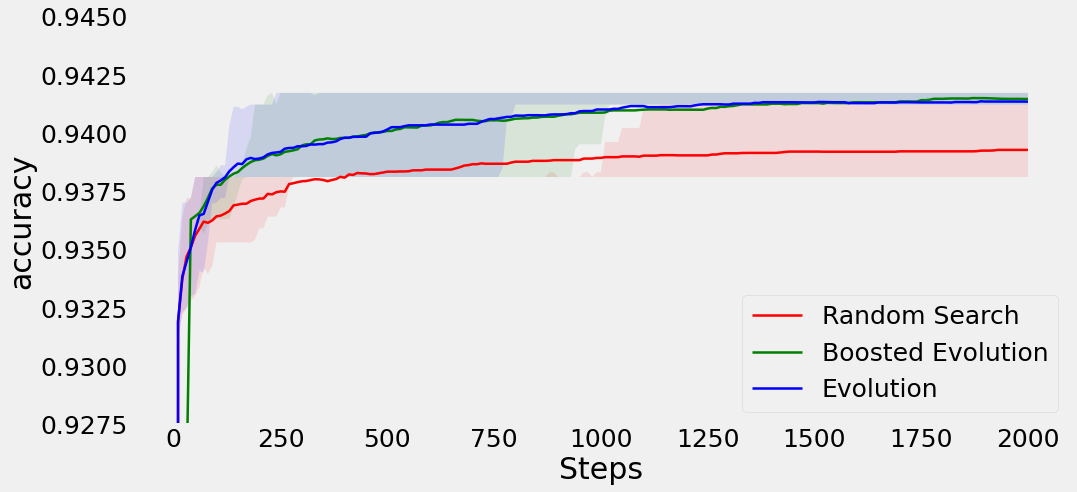}\hfill
  \caption{}
  \label{fig:boosting-ga-original-108}
  \end{subfigure}
  \caption{
  Performances in test of various NAS algorithms, when clustering with the Original Encoding.
  }
  \label{fig:boosting-ga-original}
\end{figure}

Figure~\ref{fig:boosting-ga-original} shows performances in test of various NAS Algorithms.
In red appears Random Search. In blue (Evolution) is \textit{Aging Evolution} initialized with a $N=27$ randomly selected configurations. 
In green (Boosted Evolution) is the same algorithm but initialized with the $N=27$ centroids collected from clustering. Both settings of the \textit{Aging Evolution} use a population of fixed size $P=N$, and a tournament size $S=10$. The clustering is done with BGM using the Short Encoding.
Each algorithm is ran $M=100$ independent times, each with distinct random seeds. The top (resp. bottom) 
figure is for selecting models trained for 36 (resp 108) epochs.

When selecting models at both 36 or 108 epochs, the \textit{Aging Evolution} baseline 
displays superior results with the centroid-based initialization approach using the Short encoding.
It is indeed better than the Random search baseline, and demonstrates faster convergence rates then other evaluated algorithms at 36 epochs of training.

Table~\ref{tab:perf-nas-original} summarizes the performance of the algorithms benchmarked in Figure~\ref{fig:boosting-ga-original}, i.e when clustering with the Original Encoding. 
RS stands for Random Search, AE for Aging Evolution with randomly initialized population, and B-AE for Aging Evolution with centroid-based population initialization. To complement these results, we performed a Wilcoxon rank-sum test to compare RS against B-AE. The \emph{p}-value on 36 epochs is equal to 0.0002, and on 108 epochs is equal to 1.199e-13. Thus, B-AE significantly improves over RS. 
When repeating the test with B-AE and AE, the \emph{p}-value is equal to 0.034 and 0.212 for 36 and 108 epochs respectively. We conclude that B-AE significantly improves over AE for 36 epochs.

\begin{table}[h]
    \footnotesize
    \centering
    \begin{tabular}{llrrr}
    \hline
        &    & RS & AE & B-AE \\
    \hline
    \multirow{5}{*}{36 epochs} 
    & mean   & 92.34  & 92.38  & 92.41 \\
    & median &92.33 & 92.33 & 92.33 \\
    & min & 91.98 & 92.25 &  92.18\\
    & max & 92.68 & 92.75 & 92.75\\
    & std    &  0.167 &  0.131 &  0.128 \\
    \hline
    \multirow{5}{*}{108 epochs} 
    & mean   & 93.93  & 94.13  & 94.14 \\
    & median & 93.81 & 94.12 & 94.13\\
    & min & 93.80 & 93.70 & 93.70\\
    & max & 94.32 & 94.35 & 94.42\\
    & std    &  0.164 &  0.098 & 0.132 \\
    \hline
    \end{tabular}
    \caption{Benchmark of NAS algorithm performances after 2000 iterations. The data-driven initialization technique involve the Original Encoding.}
\label{tab:perf-nas-original}
\end{table}

\begin{figure}[]
  \begin{subfigure}{\linewidth}
  \includegraphics[width=0.99\linewidth]{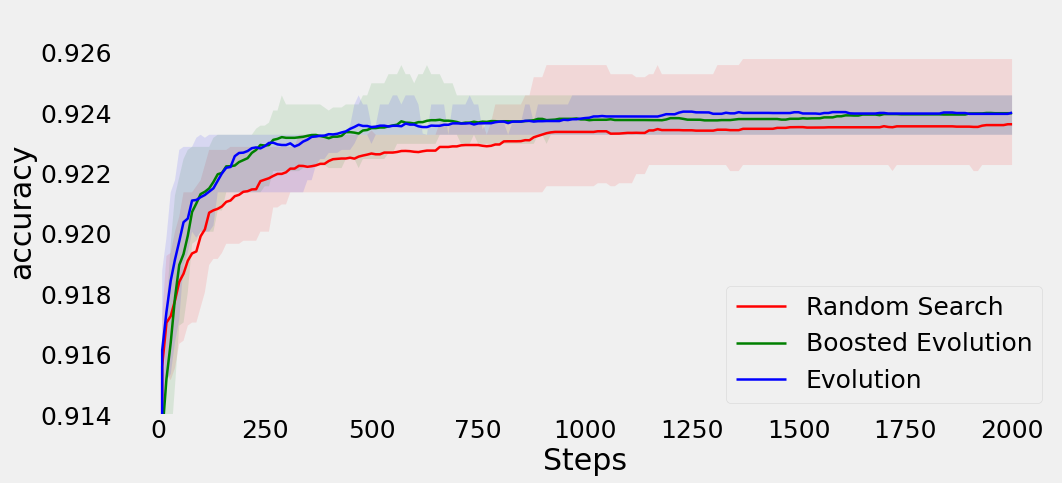}\hfill
  \caption{}
  \end{subfigure}\par\medskip
  \begin{subfigure}{\linewidth}
  \includegraphics[width=.99\linewidth]{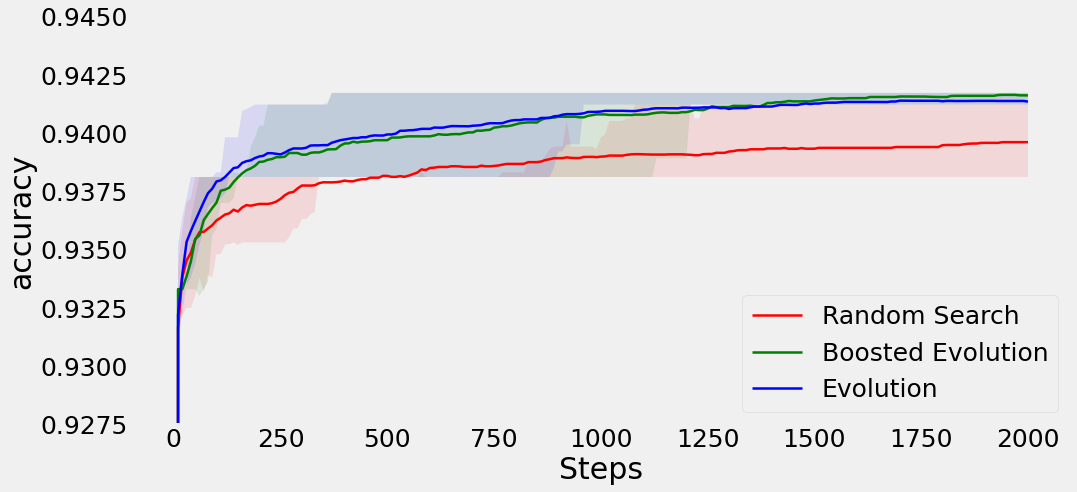}\hfill
  \caption{}
  \end{subfigure}
  \caption{
  Performances in test of various NAS algorithms, when clustering with the Binary Encoding.
  }
  \label{fig:boosting-ga-binary}
\end{figure}

Figure~\ref{fig:boosting-ga-binary} shows results for the same setting with the exception of initializing \textit{Aging Evolution} with the $N=27$ centroids collected by BGM on the Long Encoding.
Similarly, the \textit{Aging Evolution} algorithm shows competitive performances with the centroid-based initialization, improving over the baselines when training for long (108).

Table~\ref{tab:perf-nas-binary} summarizes the benchmark provided in Figure~\ref{fig:boosting-ga-binary}. 
When comparing final performances of RS and B-AE, we obtain
a \emph{p}-value at 36 epochs equal to  0.0435, and at 108 epochs equal to 5.105e-11. Thus, B-AE significantly improves over RS. 
When repeating the test with B-AE and AE, the \emph{p}-value is equal to 0.766 and  0.085 for 36 and 108 epochs respectively. 
Thus, despite higher mean and median of test accuracy at 108 epochs, the improvements of B-AE over AE are non-significant.


To summarize, the centroid extracted from a fitness-based clustering of a the Search Space 
seem to be a promising population to initializing an evolutionary search algorithm. 
In fact, in the challenging scenario of selecting models after only 36 epochs of training,
using the proposed technique with the Short Encoding improves significantly over the baseline initialization (\emph{p}-value=0.034). 

The limited improvements when clustering with the Binary Encoding 
might be explained by the fact that the \textit{Aging Evolution} is
deployed on models using the Short Encoding. 
Indeed, experiments using the Long Encoding were discarded 
because of the increased complexity for the search procedure.  
Future work might exploring this option,
at it could help better exploit the extracted population.


\begin{table}[h]
    \footnotesize
    \centering
    \begin{tabular}{llrrr}
    \hline
        &    & RS & AE & B-AE \\
    \hline
    \multirow{5}{*}{36 epochs} 
    & mean   & 92.36 & 92.40& 92.40\\
    & median & 92.33&  92.33&  92.33\\
    & min  & 91.98 &  92.25 & 92.14 \\
    & max  & 92.75&  92.75&  92.75\\
    & std  & 0.174& 0.138& 0.141\\
    \hline
    \multirow{5}{*}{108 epochs} 
    & mean  & 93.96 & 94.13& 94.16\\
    & median & 93.81&  94.12&  94.17\\
    & min & 93.70&  93.80&  93.70\\
    & max & 94.42&  94.42&  94.42\\
    & std  & 0.191& 0.118& 0.128\\
    \hline
    \end{tabular}
    \caption{Benchmark of NAS algorithm performances after 2000 iterations. The data-driven initialization technique involve the Binary Encoding.}
\label{tab:perf-nas-binary}
\end{table}


\section{Conclusion}\label{sec:conclusion}

In this study, we seek to gain insights about a search space of image classification models
in order to improve the performance of NAS algorithms.
More precisely, we want to know if the convergence of a search strategy could be improved 
using a data-driven initialization technique exploiting the search space.

For this purpose, we propose a two-step approach to improve the performances of a NAS search strategy.
First, we perform a clustering analysis of the search space, involving a sequence of sub-tasks.
It summarizes as follows: we sample models from a search space, reduce their dimension, perform a clustering.  
After a careful tuning of the clustering pipeline (number of dimensions, clusters, etc), 
we select the algorithm providing the best qualitative and quantitative results.
Second, we run a evolutionary algorithm using as an initial population the centroid extracted previously.
Results show that centroids extracted using BGM for clustering are a promising population to initialize a state-of-the-art evolutionary algorithm. 
In the scenario of selecting models trained only 36 epochs, this appoach shows faster convergence and long term improvements over a standard initialization (p-value=0.034 at 2000 iterations), when using a Short Encoding.



As future work, we propose to investigate performances of this approach when selecting models on the Long Encoding. We also propose to study in depth the obtained clusters to gain more insights on obtained performances. One might also explore the benefits of such data-driven initialization method on other families of algorithms (Bayesian Optimization, Local Search, etc).

\vspace{-0.35cm}
\section*{Acknowledgments}\label{ACKNOWLEDGEMENTS}

Authors acknowledge support by the European Research Council (ERC) under the European Union's Horizon 2020 research and innovation program (grant agreement No. [ERC-2016-StG-714087], Acronym: \textit{So2Sat}), by the Helmholtz Association
through the Framework of Helmholtz AI [grant  number:  ZT-I-PF-5-01] - Local Unit ``Munich Unit @Aeronautics, Space and Transport (MASTr)'' and Helmholtz Excellent Professorship ``Data Science in Earth Observation - Big Data Fusion for Urban Research''(W2-W3-100),  by the German Federal Ministry of Education and Research (BMBF) in the framework of the international future AI lab "AI4EO -- Artificial Intelligence for Earth Observation: Reasoning, Uncertainties, Ethics and Beyond" (Grant number: 01DD20001). and the grant DeToL

\bibliographystyle{named}
\bibliography{ijcai21}

\end{document}